\documentclass{article}

\usepackage[preprint]{arxiv}

\usepackage[utf8]{inputenc} 
\usepackage[T1]{fontenc}    
\usepackage{hyperref}       
\usepackage{url}            
\usepackage{booktabs}       
\usepackage{amsfonts}       
\usepackage{nicefrac}       
\usepackage{microtype}      
\usepackage{xcolor}         
\usepackage{graphicx}
\usepackage{amsmath}
\usepackage{multirow}

\title{Building Decision Making Models Through Language Model Regime}

\author{%
  Yu Zhang \hspace{1em} 
  Haoxiang Liu \hspace{1em} 
  Feijun Jiang \hspace{1em} 
  Weihua Luo \hspace{1em} 
  Kaifu Zhang \\ \\
  \vspace{1em} 
  \textit{Alibaba Group} \\
  \vspace{1em} 
  \texttt{\{yuni.zy, haoxiang.lhx, feijun.jiangfj, weihua.luowh, kaifu.zkf\}@alibaba-inc.com}
}

\begin{document}

\maketitle

\begin{abstract}
  We propose a novel approach for decision making problems leveraging the generalization capabilities of large language models (LLMs). Traditional methods such as expert systems, planning algorithms, and reinforcement learning often exhibit limited generalization, typically requiring the training of new models for each unique task. In contrast, LLMs demonstrate remarkable success in generalizing across varied language tasks, inspiring a new strategy for training decision making models. Our approach, referred to as "Learning then Using" (LTU), entails a two-stage process. Initially, the \textit{learning} phase develops a robust foundational decision making model by integrating diverse knowledge from various domains and decision making contexts. The subsequent \textit{using} phase refines this foundation model for specific decision making scenarios. Distinct from other studies that employ LLMs for decision making through supervised learning, our LTU method embraces a versatile training methodology that combines broad pre-training with targeted fine-tuning. Experiments in e-commerce domains such as advertising and search optimization have shown that LTU approach outperforms traditional supervised learning regimes in decision making capabilities and generalization. The LTU approach is the first practical training architecture for both single-step and multi-step decision making tasks combined with LLMs, which can be applied beyond game and robot domains. It provides a robust and adaptable framework for decision making, enhances the effectiveness and flexibility of various systems in tackling various challenges.  
\end{abstract}

\section{Introduction}
The decision making problem has consistently presented as a significant challenge in the field of deep learning \citep{Phillips-Wren}, driving researchers and practitioners to explore innovative solutions. Traditionally, methods such as expert systems \citep{Luger1990ArtificialIA,Chaudhary,Sharipbay}, planning algorithms \citep{Ortolano1984EnvironmentalPA,Krantz2007GoalsAP,Sousa2015PlanningTD,Kasie2017DecisionSS,Liu2021DecisionMakingTF}, as well as reinforcement learning techniques \citep{Abel2016ReinforcementLA,Aradi2020SurveyOD},have been fundamental approaches for tackling complex decision making tasks. 

However,  these methods are all facing poor generalization abilities \citep{Pannu2015SurveyOE,Reed2022AGA}, although they work well within their specific applications. Using these methods for new or unexpected tasks are not straightforward and usually require developing and training completely new models for each task. Yet, the inherent similarities are undoubtedly existing across decision making tasks with similar background, like computer games of same types  share a wide range of  common skills \citep{Reed2022AGA}. Traditional strategies have struggled to identify and use these commonalities to create a decision making model that is both general and flexible. 


The rising of large language models (LLM) \citep{Radford2019LanguageMA,Brown2020LanguageMA,openai2024gpt4} has changed what might be possible in this area. With massive data and model parameters build on transformer architecture \citep{Vaswani2017AttentionIA}, large language models have been very successful at generalizing across a wide range of language tasks \citep{openai2024gpt4}. Their ability to effectively deal with all kinds of natural language understanding (NLU) and natural language generation (NLG) problems provide new ideas for enhancing the generalization abilities of decision making models. Recent works explored how to use paradigm of language modeling to solve decision making problems \citep{Chen2021DecisionTR,Janner2021OfflineRL}, but they have largely focused on sequence modeling with the paradigm of supervised tuning directly. However, the generalization and emergent abilities of large language models are not a product of supervised learning on specific tasks \citep{Wei2022EmergentAO}. Rather, they are achieved through self-supervised pre-training, learning from vast amounts of data that are not entirely aligned with any specific tasks \citep{openai2024gpt4,Claude3}. But ultimately, these models are capable of tackling specific downstream tasks and perform exceedingly well at them. 

Inspired by how large language model is trained and used, we propose a new method to develop a decision making model that goes beyond the limitations of task-specific models. This method seeks to improve both effectiveness and adaptability with which decision making tasks can be approached. Unlike supervised tuning paradigm, where models are trained in a supervised way and then are expected to encounter similar data and make correct decisions during application scenarios \citep{Ouyang2022TrainingLM}, we introduce a new training strategy called "\textit{Learning then Using}" (\textbf{LTU}). This is a two-step method starts with a continue pre-training (CT) phase \citep{Gupta2023ContinualPO}, which aims to build a foundation decision making model via infusing a wide range of knowledge across various related decision domains and contexts. This phase is responsible for \textit{learning}. Following this, the supervised fine-tuning (SFT) phase further refines the foundation model to reach requirements of a certain decision making task. The SFT phase is responsible for \textit{using}, aiming to solve specific downstream tasks.

By combining the broad potential of large language models with a carefully designed training process, we study whether this new paradigm can provide better decision making abilities than supervised learning and exhibit stronger generalization. We conduct experiments in two classic e-commerce scenarios, advertising and searching. Across a total of four tasks with various ablation studies, LTU method not only outperforms SFT on decision making tasks but also makes the foundation model built with \textit{learning} phase remarkably versatile. When applied to different decision making tasks, the foundation model consistently performs better than those trained only through supervised learning regimes. 

\section{Related work}
decision making tasks combined with language modeling has notably impacted the decision making domain in recent studies \citep{Yang2023FoundationMF,Jiang2023LargeLM,Wen2022MultiAgentRL}. There is a growing interest in building foundation decision making models to solve multiple decision tasks in one model and to easily transfer to new downstream tasks without training from  scratch \citep{Wen2022OnRO,10.24963/ijcai.2023/808,Meng2021OfflinePM}.  Studies have explored the feasibility of building foundation models for decision making based on language models \citep{Yang2023FoundationMF,Jiang2023LargeLM} and discussed various topics of combining language models with decision making tasks. Some other studies \citep{10.24963/ijcai.2023/808} work to offer a explicit definition of  foundation decision making models and theoretically possible training paradigms.

Besides the research on foundation models, which largely remains at the theoretical stages, many efforts have already been made to solve decision problems by practically applying language modeling methods. Decision Transformer \citep{Chen2021DecisionTR} leverages transformer architecture to learn offline reinforcement learning tasks as conditioned generating problems. It requires both expected reward and current environment state as the input to transformer, aiming to prompt the model to output corresponding actions. Remarkably,  Decision Transformer has demonstrated superior performance compared to offline RL methods in relevant tasks. Trajectory Transformer \citep{Janner2021OfflineRL} . suggests to build whole transactions of reinforcement learning (\textit{state}, \textit{action}, and \textit{reward}) into a sequential modeling learning paradigm  and employed beam search for action selection.  The authors argue that it simplifies a range of design decisions and eliminates the need for many components typically found in offline RL algorithms.

In language model domains, efforts have been made to combine pre-training with Supervised Fine-Tuning (SFT) to solve specify downstream tasks. \citep{Askell2021AGL} proposes Preference Model Pre-training(PMP) to train a preference model (also known as reward model). It first utilize millions of sentence pairs with specific format to pre-train a large language model to get a 'foundation preference model'. Then it performs a SFT with pair data based on the foundation model to get a task related preference model. Ablation studies reveal that preference models trained with PMP surpass those trained with only SFT using task-related pair data in terms of performance.   Additionally, \citep{AllenZhu2024PhysicsOL} organizes pre-training data into tuples \textit{(name, attribute, value) }to perform pre-training and later tests the models on downstream tasks to check whether large language models can effectively scale up when equipped with structured pre-training data.

Our work aims to extend decision making models beyond sequential tasks like gaming and robotics and propose a novel method that moves away from direct supervised learning, offering a method of how to build a foundation decision making models and then use it in specific tasks.


\section{Methodology}
\label{sec:Methodology}
\subsection{Preliminaries}
Our methodology employs Llama-2-13b \citep{ Touvron2023Llama2O} , a pre-trained large language model (LLM), as base model for training. We adopt the causal language modeling (CLM) training paradigm and categorize our data based on a decision making pattern with the transformer architecture \citep{Janner2021OfflineRL}.

\textbf{Causal Language Modeling Approach:} Consider a sequence $x = [x_1, x_2, \ldots, x_{|x|}]$ of variable length $|x|$, which could represent a piece of text or a snippet of code. The task of Causal Language Modeling (CLM) primarily revolves around predicting the distribution of a series of sequences $x^1, x^2, \ldots, x^N$. To accurately approximate this distribution, a prevalent strategy is to break down the joint probability distribution of each sequence into a series of probabilities predicting each subsequent token given the prior ones. The optimization of the model is achieved by utilizing the principle of maximum likelihood estimation as presented in Eq\ref{eq:clm} \citep{Jain2022ContraCLMCL},

\begin{equation}
L_{\text{CLM}} = -\frac{1}{N} \sum_{j=1}^{N} \sum_{i=1}^{|x^j|} \log p(x_i^j | x_{<i}^j) \label{eq:clm}
\end{equation}

where \(x_{<i}^j=[x_{1}^j, \ldots, x_{i-1}^j]\) represents the subsequence preceding \(x_{i}^j\), and \(|x^j|\) indicates the sequence length.

\begin{figure}
    \centering
    \includegraphics[width=0.75\linewidth]{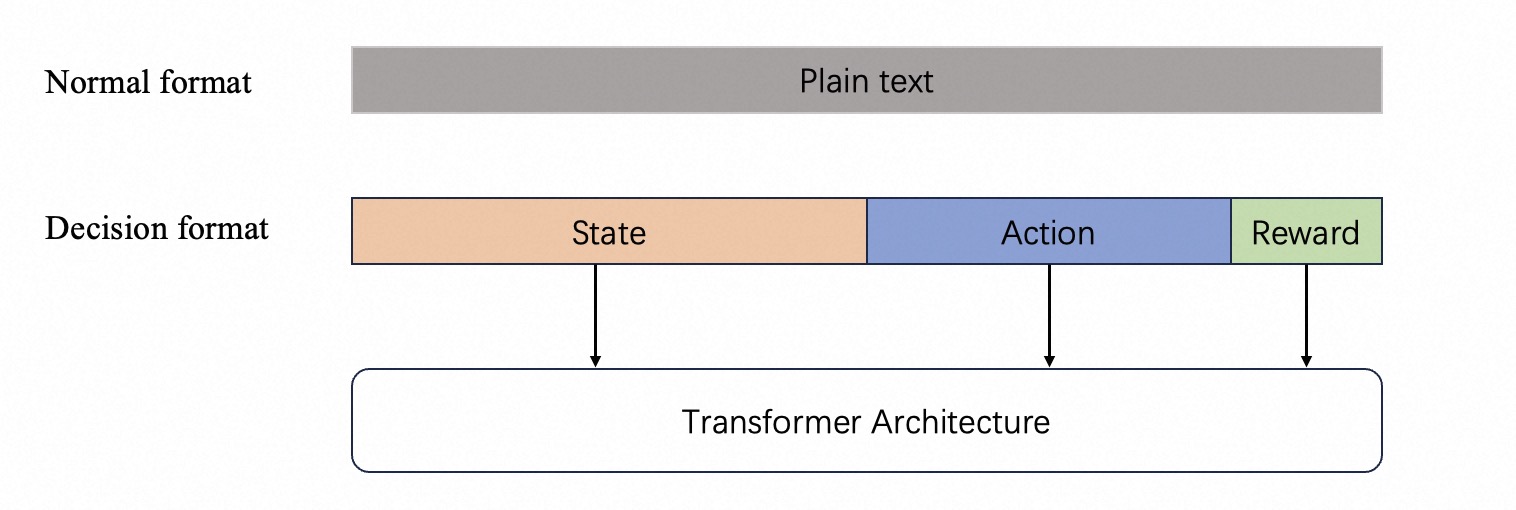}
    \caption{Different context part in transformer architecture}
    \label{fig:transformer-context}
\end{figure}

\textbf{Components of decision making:} While causal language modeling aims to predict the next token and maximize the probability of tokens during training, it does not explicitly consider the different part of context. Recent studies have demonstrated the critical role of the patterns implicit in contexts or sequences for decision making in transformer architectures \citep{Chen2021DecisionTR,Janner2021OfflineRL,Gunasekar2023TextbooksAA}. Our approach classifies all contexts in our training data into three components: $S$, $A$, and $R$ . $S$ represents the \textit{state}, encompassing the decision making background, task-related details, and the information necessary for making decisions. This notion of \textit{state} shares the concept in reinforcement learning, where it typically means the complete elements of current environment for making decisions. $A$ denotes the \textit{action}, which is the specific measure taken given the \textit{state}. This could be a simulated action in a game, like playing Atari, or a real-world action, such as taking a bus or purchasing a book. In the context of this study, actions are transformed into text format. $R$ stands for \textit{reward}, the feedback received following the actions taken in specific states. The reward could be a quantitative measure like a user rating, or qualitative feedback such as 'I love it' or 'She is happy with it'. Figure \ref{fig:transformer-context} illustrates the basic pattern of $S, A, R$ within a transformer architecture.

\subsection{Data Construction}
\label{sec:Data Construction}
Constructing training data for our proposed method can be approached in various ways, including formats like $(S,A)$, $(S,R,A)$, $(S,A,R)$, and $(R, S, A)$. Determining the most effective data structure for training such decision models is still an open problem \citep{Wen2022MultiAgentRL}. In our approach, we organize our data in $(s,a,r)$ form. The length of  a single data depends on different decision tasks. For single step decision problems, it can just be $(s,a,r)$, and for sequential decision tasks, it can be formulated as $(s_1, a_1, r_1, s_2, a_2, r_2, ...s_n, a_n, r_n)$. This data formulation let the model learn to  functioning as a reward model or evaluator because the majority of decision making problems can be build as either rating issues or reward-based selection dilemmas.  

As for the \textit{using} phase, we construct  text before the final $R$ as an input, and takes final $R$ as output. For single step decision problems, it is $(s,a)$ and $r$, and for sequential decision problems, it is $(s_1, a_1, r_1, s_2, a_2, r_2, ... s_n, a_n)$ and $r_n$.

\subsection{Training Paradigm}
\label{sec:Training Paradigm}
The conventional method for aligning language models with downstream tasks, such as decision making, employs supervised fine-tuning (SFT). We introduce a training paradigm named \textbf{Learning then Using}(LTU), with separate \textit{learning} and \textit{using} phase.

\textbf{Learning in LTU}. Our approach involves an initial \textit{learning} phase with continued pre-training on large language models (LLMs). Continue pre-training(CT) is widely used to enhence abilities of different domains or languages based on already well-trained large language models \citep{Cui2023EfficientAE, Rozire2023CodeLO}. The basic idea of LTU method is LLMs are able to learn inherent patterns and statistics correlations via decision making knowledge formulated as $(s, a, r)$. Through continued pre-training, we can integrate this collective decision making intelligence into a LLM , transforming it into a comprehensive foundation decision making model which is suitable for various downstream tasks.  We construct the data format in $(s,a,r)$ pairs as we mentioned in \ref{sec:Data Construction}. We use this part of data to do an auto-regression training  following the Eq \ref{eq:clm} based on a LLM.  After the \textit{learning} phase, we get our foundation decision making model.


\textbf{Using in LTU.} The \textit{using} phase is a classic supervised fine-tuning phase. Supervised fine-tuning(SFT) has emerged as a powerful and effective technique to adapt pre-trained LLMs  to specific downstream tasks through supervised learning. In this part, we leverage the foundation decision making model trained in \textit{learning} phase and learn to predict $P(r|s,a)$ to solve certain decision making tasks.  


\section{Experiments}
In this section, we are aiming to answer the following questions:
\begin{enumerate}
    \item Does LTU outperform supervised fine-tuning in decision making tasks?
    \item Do models trained with LTU have stronger generalization capabilities compared to supervised fine-tuning?
    \item How does incorporating common knowledge data influence the LTU training process?
\end{enumerate}

\subsection{Training Setup}
Our experiments are conducted across two e-commerce-related tasks: a pay-per-click(PPC) advertisement task and a search-engine-optimization(SEO) task. For the PPC task,  our objective is to predict click-through rate (CTR)  and cost-per-click (CPC)  for a certain advertisement.  As for the SEO task, our objective is to predict impressions of products and its click-through rate (CTR). For each ablation experiment, we train one foundation model and four separate SFT models by LTU method. 

\begin{figure}[h]
    \centering
    \includegraphics[width=1\linewidth]{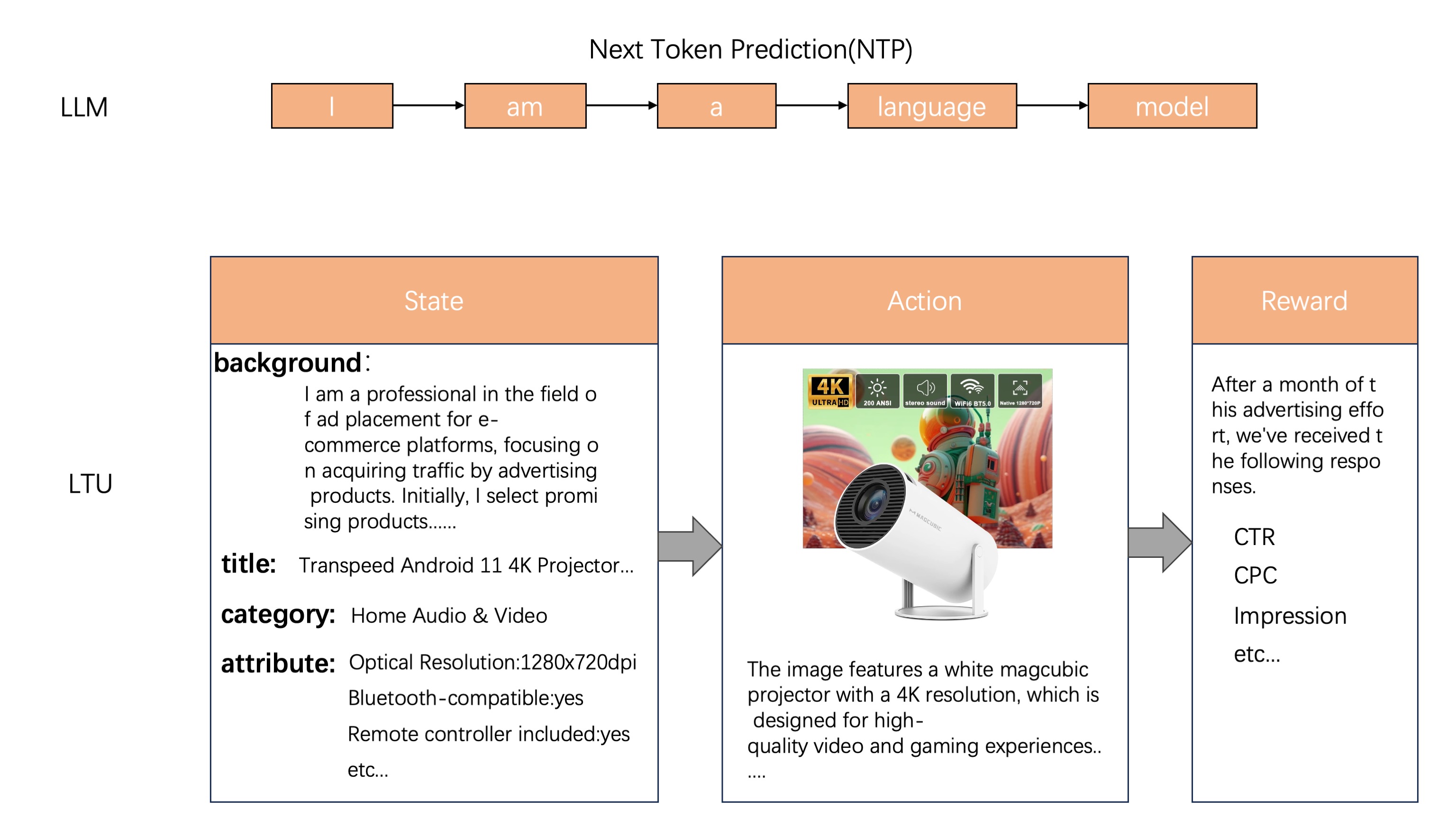}
    \caption{Example of PPC training data}
    \label{fig:PPC-data}
\end{figure}

We collect our datasets from e-commerce platforms, gathering online data with genuine impressions and user click activities. We collect our SEO data from AliExpress.com and our PPC data from Facebook advertisement system. Training data is all turned into $(s,a,r)$ format. In both  PPC and SEO tasks, $s$ represents product information, including product titles, attributes, and detailed descriptions. $a$ represents product pictures made to publish as advertising images for PPC and as main pictures showed at search result page for SEO.  And $r$ corresponds to task-specific metrics such as CTR, CPC, or impressions, all of which are discretized.  Figure \ref{fig:PPC-data} \ illustrates a simple $(s,a,r)$  PPC data flow.  For PPC task, reward values such as CTR and CPC are categorized into three groups, ranging from 0 to 2. For SEO task, CTR and impressions are categorized into ten groups, from 0 to 9.  We collect 23340791 search results for SEO task and 318766 advertising results for PPC task. 

We conduct LTU training on these two datasets. In \textit{learning} phase, we  train totally four foundation decision making models through continue pre-training for ablation studies and each of them is trained with 10 billion tokens on 64 A100 GPUs.  All these four models are trained based on Llama-2-13b \citep{ Touvron2023Llama2O}. With SEO data, which contains 19.562 billion tokens in total, we sampled 10 billion tokens to train a model named \textbf{LTU-SEO}. For PPC data, which contains only 0.22 billion tokens, we supplement it with general open-source data such as RefinedWeb, Wudao to reach a total of 10 billion training tokens and train a foundation model named \textbf{LTU-PPC}. Furthermore, for SEO data, we train another model, \textbf{LTU-SEO-MIX}, integrating SEO data with common knowledge to study data-mixing impact on LTU training. Additionally, we trained a model with 80 billion tokens of common knowledge which are irrelevant with decision making referred to as \textbf{LTU-common}. The detailed composition of training data for each model is in Table~\ref{tab:ct-data}. 
\begin{table}[h]
    \centering
    \caption{Detailed CT training data}
    \label{tab:ct-data}
    \renewcommand{\arraystretch}{1.2} 
    \begin{tabular}{|c|c|c|c|c|} 
        \hline
         &    LTU-SEO & LTU-SEO-MIX & LTU-PPC & LTU-Common \\ \hline 
         Advertise data &    0B & 0B & 0.22B & 0B \\ \hline 
         Search data &   10B & 2.5B & 0B & 0B \\ \hline 
         Common Knowledge &   0B & 7.5B & 9.78B & 80B \\ \hline
    \end{tabular}
\end{table}

Followed by \textit{learning} phase with four foundation models trained, we begin to study whether these models can perform better in \textit{using} phase compared with supervised fine-tuning.

\textbf{Compare decision making abilities of LTU and Supervised Fine-tuning.}In this part, we study decision making abilities through PPC task. For LTU method, we perform \textit{using} part on \textbf{LTU-PPC} model, which makes it an in-domain \textit{learning} and \textit{using} task. For  supervised fine-tuning(SFT), we perform supervised learning directly through Llama-2-13B. We test accuracy of CTR and CPC prediction in PPC task. We construct a dataset of 25,009 PPC-specific SFT samples, designated as \textit{PPC-SFT}, which is excluded from the CT training phase. Additionally, we created a SFT dataset named \textit{PPC-SFT-full} by transforming our CT training data into SFT format and merging it with \textit{PPC-SFT}. This resulted in a total of 343775 samples in the \textit{PPC-SFT-full} dataset.

We train and evaluate models using both \textit{PPC-SFT} and \textit{PPC-SFT-full} datasets. As illustrated in Table~\ref{tab:LTU-PPC}, LTU models demonstrate superior accuracy compared to SFT model. For the CTR prediction task, the LTU method outperforms the SFT across both PPC datasets. Specifically, in the \textit{PPC-SFT} data, the LTU method achieved an accuracy of 0.621 compared to the SFT's 0.599, highlighting its superior ability to predict user engagement with ads. Similar trends are observed in the prediction of CPC. The LTU method again demonstrates enhanced accuracy over SFT on both PPC datasets. In the \textit{PPC-SFT} context, LTU's accuracy is 0.572, outpacing the SFT's 0.548 and suggesting more precise prediction capabilities for advertising costs. To illustrate that improvements are due to decision making data rather than an infusion of common knowledge, we compare the LTU-PPC results with LTU-Common model, which is trained solely on common knowledge. The results, presented in Table \ref{tab:LTU-PPC} as \textit{LTU-Common Accuracy}, reveal that LTU-Common model significantly underperform the LTU-trained models and even fall short of the SFT model, highlighting a potential issue of knowledge forgetting and prove that decision making related knowledge learned during \textit{learning} phase contribute most to downstream decision making tasks.

\begin{table}[h]
    \centering
\caption {Results on PPC Tasks}
\label{tab:LTU-PPC}
\renewcommand{\arraystretch}{1.2} 
    \begin{tabular}{|c|c|c|c|c|} \hline
         Task & Data & LTU-PPC Accuracy& SFT Accuracy& LTU-Common Accuracy\\ \hline
         \multirow{2}{*}{Predict CTR}& \textit{PPC-SFT} & 0.621 & 0.599 & - \\ \cline{2-5}
         & \textit{PPC-SFT-full} & 0.646 & 0.634 & 0.591 \\ \hline
         \multirow{2}{*}{Predict CPC}& \textit{PPC-SFT} & 0.572 & 0.548 & - \\ \cline{2-5}
          & \textit{PPC-SFT-full} & 0.616 & 0.598 & 0.564 \\ \hline
    \end{tabular}
\end{table}

\textbf{Comparing Generalization Abilities of LTU and Supervised Fine-tuning.} In this part, we focus on generalization capabilities of LTU and supervised fine-tuning (SFT) models. For LTU method, we utilize LTU-SEO as foundation model for downstream experiments, a model trained only on SEO related data. And for SFT method, we still perform supervised learning directly through Llama-2-13B. 

We use two datasets in this part for training and evaluation. The first one is \textit{SEO-SFT}. It consists of two categories of SEO data, both of which are not included in the training data of LTU-SEO. As a result, both LTU-SEO and Llama-2-13B have never seen these two types of categories before. We use this dataset to study trans-category generalization ability in LTU and SFT methods since strategies can vary significantly across different categories in SEO scenarios. The second dataset we use in this part is \textit{PPC-SFT-full}, the dataset we have built to study  decision making abilities on PPC tasks. It consists of PPC data from Fackbook advertisement system, which is completely different from SEO data that collected from AliExpress.com. We use this dataset to compare the out-of-domain generalization abilities between LTU and SFT methods. 

We first train on CTR prediction and impression prediction tasks of SEO. Based on the presented result in Table \ref{tab:SEO-trans-cate}, we observe that LTU method outperforms the SFT method in terms of accuracy for both CTR and impression prediction tasks within the context of SEO. The advantages suggest that LTU method have better capabilities in predicting behavior in unseen SEO categories, which indicates a stronger generalization capability for in-domain decision making tasks.

\begin{table}[h]
    \centering
\caption{Results on SEO tasks }
\label{tab:SEO-trans-cate}
\renewcommand{\arraystretch}{1.2} 
    \begin{tabular}{|c|c|c|c|} \hline 
         Task & Data & LTU-SEO Accuracy& SFT Accuracy\\ \hline 
         CTR & \textit{SEO-SFT}&  0.156 & 0.148 \\ \hline 
         Impression & \textit{SEO-SFT}& 0.146 & 0.126 \\ \hline
    \end{tabular}
\end{table}


We further study the performance of LTU method on out-of-domain tasks by using \textit{PPC-SFT-full} and trained from LTU-SEO model. We conduct CPC and CTR prediction learning tasks and evaluate the accuracy on test sets. The LTU method once again demonstrate enhanced performance in both PPC tasks, which are from significantly different domains. The results in Table \ref{tab:seo-result} show that accuracy on CPC task is 3.0\% higher and on CTR task is 3.01\% higher than SFT. 

\begin{table}[h]
    \centering
\caption{PPC results on LTU-SEO and SFT}
\label{tab:seo-result}
\renewcommand{\arraystretch}{1.2} 
    \begin{tabular}{|c|c|c|c|} \hline 
         Task&  Data&  LTU-SEO Accuracy& SFT Accuracy\\ \hline 
         CPC&  \textit{PPC-SFT-full} &  0.653&  0.634\\ \hline 
         CTR&  \textit{PPC-SFT-full} &  0.616&  0.598\\ \hline
    \end{tabular}

\end{table}

Additionally, we investigate weather the superior performance of LTU is a result of novel methodology or is just a result of engaging more training data during \textit{learning} phase compared with SFT. We convert all our SEO data used for training LTU-SEO in \textit{learning} phase into supervised learning style and combine it with \textit{PPC-SFT-full} dataset, getting a new dataset \textit{SFT-full}. We utilize this dataset to train SFT model and compare it with LTU method trained with LTU-SEO model and \textit{PPC-SFT-full} dataset, where it shares exactly same amount of data with SFT.  Comparison results in Table \ref{tab:seo-sft-full-result} show LTU outperfomers SFT trained with \textit{SFT-full} dataset about 1.90\% on CPC task and  2.84\% on CTR task, suggesting that LTU is a novel methodology that has more data efficiency and generalization abilities compared with SFT. 

\begin{table}[h]
    \centering
    \caption{PPC results on same size of training  data }
    \label{tab:seo-sft-full-result}
    \renewcommand{\arraystretch}{1.2} 
    \begin{tabular}{cccc} 
        \hline
        \textbf{Task} & \textbf{Data} & \textbf{Method} & \textbf{Accuracy} \\
        \hline
        \multirow{2}{*}{CPC} & \textit{PPC-SFT-full} & LTU-SEO & 0.653\\
        \cline{2-4}
                             & \textit{SFT-full} & SFT & 0.641 \\
        \hline
        \multirow{2}{*}{CTR} & \textit{PPC-SFT-full} & LTU-SEO & 0.616 \\
        \cline{2-4}
                             & \textit{SFT-full} & SFT & 0.599 \\
        \hline
    \end{tabular}
\end{table}

\textbf{Impact of Common Knowledge on Training Effectiveness}. In this part, we aim to investigate the impact of integrating common knowledge into training process of LTU method. The LTU-SEO model is initially trained on a dataset comprising 10 billion tokens without incorporating any common knowledge. We develope a hybrid model named LTU-SEO-MIX, which is trained using a combination of 2.5 billion tokens of SEO data and 7.5 billion tokens of common knowledge. We evaluate the performance of both models on SEO and PPC tasks.

The results, as presented in Table \ref{tab:ltu-common}, show that the inclusion of common knowledge data negatively affects the training of LTU method. The LTU-SEO-MIX model exhibit inferior performance on both SEO and PPC tasks compared to its counterpart trained exclusively on SEO data. This result aligns with research from training of code LLMs where it has been reported that a dataset consisting solely of code data outperforms a mixed dataset of common knowledge \citep{Rozire2023CodeLO}. 

\begin{table}[h]
    \centering
\caption{SEO and PPC results on models with different common knowledge }
\label{tab:ltu-common}
\renewcommand{\arraystretch}{1.2} 
    \begin{tabular}{|c|c|c|c|} \hline 
         Task&  Data&  LTU-SEO Accuracy& LTU-SEO-MIX Accuracy\\ \hline 
          CTR&  \textit{SEO-SFT}&  0.163& 0.156\\ \hline 
         Impression&  \textit{SEO-SFT}&  0.153& 0.149\\ \hline
  CTR& \textit{PPC-SFT-full}& 0.653&0.645\\\hline
 CPC& \textit{PPC-SFT-full}& 0.616&0.616\\\hline
    \end{tabular}
    
\end{table}

\section{Conclusion}
In this paper, we present a novel approach of building decision making models called "Learning then Using" (LTU), which bridges the generalization capabilities of large language models (LLMs) with decision making tasks. Our approach involves a two-step process consisting of building a foundation decision making model through continued pre-training (CT) and use it on downstream tasks via supervised fine-tuning (SFT). By leveraging the broad pre-training of LLMs and fine-tuning on domain-specific data, we aim to enhance the model's decision making abilities and generalization performance.

Our experimental results demonstrate that the LTU approach surpasses standard supervised learning paradigms in both effectiveness and adaptability in decision making tasks. Through extensive experiments on two classic e-commerce scenarios, we show that LTU  method outperforms SFT in decision making abilities and generalization capabilities.

The ablation studies suggest that LTU's superior performance is attributed to the comprehensive knowledge injected during the CT phase. We find that even for out-of-distribution data, LTU-trained models exhibit stronger generalization capabilities than SFT models. Furthermore, our analysis reveals that incorporating common knowledge into the training process can be detrimental, possibly due to knowledge forgetting or an overemphasis on broader contextual understanding at the expense of task-specific patterns.

While we believe LTU method along with the experiment result present a promising picture for decision making problems, work on this subject remains in an early stage. Our experiment is conducted on e-commercial related tasks, it still needs exploration and practical experiments on other domains. Besides, both tasks we tested are single-step decision making problems. While our proposed training architecture is fit for sequence decision making, efforts and fully conducted ablation studies are needed to prove its effectiveness. 

In conclusion, our approach to build decision making models through the language model regime opens new avenues for the development of advanced decision making systems. By combining the rich semantic understanding of LLMs with a well-curated training regimen, we offer a framework that not only enhances performance but also adapts flexibly to a variety of decision making scenarios. This strategic interplay of \textit{learning} and \textit{using} promises to deliver more robust, generalizable, and effective decision making solutions across different tasks.

\newpage
\bibliographystyle{apalike}  
\bibliography{arxiv}

\begin{thebibliography}{}

\bibitem[Abel et~al., 2016]{Abel2016ReinforcementLA}
Abel, D., MacGlashan, J., and Littman, M.~L. (2016).
\newblock Reinforcement learning as a framework for ethical decision making.
\newblock In {\em AAAI Workshop: AI, Ethics, and Society}.

\bibitem[Allen-Zhu and Li, 2024]{AllenZhu2024PhysicsOL}
Allen-Zhu, Z. and Li, Y. (2024).
\newblock Physics of language models: Part 3.3, knowledge capacity scaling laws.
\newblock {\em ArXiv}, abs/2404.05405.

\bibitem[Anthropic, 2024]{Claude3}
Anthropic (2024).
\newblock The claude 3 model family: Opus, sonnet, haiku.
\newblock {\em ArXiv}.

\bibitem[Aradi, 2020]{Aradi2020SurveyOD}
Aradi, S. (2020).
\newblock Survey of deep reinforcement learning for motion planning of autonomous vehicles.
\newblock {\em IEEE Transactions on Intelligent Transportation Systems}, 23:740--759.

\bibitem[Askell et~al., 2021]{Askell2021AGL}
Askell, A., Bai, Y., Chen, A., Drain, D., Ganguli, D., Henighan, T., Jones, A., Joseph, N., Mann, B., Dassarma, N., Elhage, N., Hatfield-Dodds, Z., Hernandez, D., Kernion, J., Ndousse, K., Olsson, C., Amodei, D., Brown, T.~B., Clark, J., McCandlish, S., Olah, C., and Kaplan, J. (2021).
\newblock A general language assistant as a laboratory for alignment.
\newblock {\em ArXiv}, abs/2112.00861.

\bibitem[Brown et~al., 2020]{Brown2020LanguageMA}
Brown, T.~B., Mann, B., Ryder, N., Subbiah, M., Kaplan, J., Dhariwal, P., Neelakantan, A., Shyam, P., Sastry, G., Askell, A., Agarwal, S., Herbert-Voss, A., Krueger, G., Henighan, T., Child, R., Ramesh, A., Ziegler, D.~M., Wu, J., Winter, C., Hesse, C., Chen, M., Sigler, E., Litwin, M., Gray, S., Chess, B., Clark, J., Berner, C., McCandlish, S., Radford, A., Sutskever, I., and Amodei, D. (2020).
\newblock Language models are few-shot learners.
\newblock {\em ArXiv}, abs/2005.14165.

\bibitem[Chaudhary et~al., 2024]{Chaudhary}
Chaudhary, J., Parmar, N., and Mehta, D. (2024).
\newblock Artificial intelligence and expert systems.
\newblock {\em International Journal of Advanced Research in Science, Communication and Technology}, pages 535--546.

\bibitem[Chen et~al., 2021]{Chen2021DecisionTR}
Chen, L., Lu, K., Rajeswaran, A., Lee, K., Grover, A., Laskin, M., Abbeel, P., Srinivas, A., and Mordatch, I. (2021).
\newblock Decision transformer: Reinforcement learning via sequence modeling.
\newblock In {\em Neural Information Processing Systems}.

\bibitem[Cui et~al., 2023]{Cui2023EfficientAE}
Cui, Y., Yang, Z., and Yao, X. (2023).
\newblock Efficient and effective text encoding for chinese llama and alpaca.
\newblock {\em ArXiv}, abs/2304.08177.

\bibitem[de~Sousa et~al., 2015]{Sousa2015PlanningTD}
de~Sousa, W.~H., Porto, M. C.~G., Marcatonio, M. I.~P., Takenouchi, P.~I., and Yu, A. S.~O. (2015).
\newblock Planning the decision making process: A multiple case study.
\newblock {\em Engineering Management Research}, 4:82--96.

\bibitem[Gunasekar et~al., 2023]{Gunasekar2023TextbooksAA}
Gunasekar, S., Zhang, Y., Aneja, J., Mendes, C. C.~T., Giorno, A.~D., Gopi, S., Javaheripi, M., Kauffmann, P.~C., de~Rosa, G., Saarikivi, O., Salim, A., Shah, S., Behl, H.~S., Wang, X., Bubeck, S., Eldan, R., Kalai, A.~T., Lee, Y.~T., and Li, Y.-F. (2023).
\newblock Textbooks are all you need.
\newblock {\em ArXiv}, abs/2306.11644.

\bibitem[Gupta et~al., 2023]{Gupta2023ContinualPO}
Gupta, K., Th'erien, B., Ibrahim, A., Richter, M.~L., Anthony, Q.~G., Belilovsky, E., Rish, I., and Lesort, T. (2023).
\newblock Continual pre-training of large language models: How to (re)warm your model?
\newblock {\em ArXiv}, abs/2308.04014.

\bibitem[Jain et~al., 2022]{Jain2022ContraCLMCL}
Jain, N., Zhang, D., Ahmad, W.~U., Wang, Z., Nan, F., Li, X., Tan, M., Nallapati, R., Ray, B., Bhatia, P., Ma, X., and Xiang, B. (2022).
\newblock Contraclm: Contrastive learning for causal language model.
\newblock In {\em Annual Meeting of the Association for Computational Linguistics}.

\bibitem[Janner et~al., 2021]{Janner2021OfflineRL}
Janner, M., Li, Q., and Levine, S. (2021).
\newblock Offline reinforcement learning as one big sequence modeling problem.
\newblock In {\em Neural Information Processing Systems}.

\bibitem[Jiang et~al., 2023]{Jiang2023LargeLM}
Jiang, H., Ge, L., Gao, Y., Wang, J., and Song, R. (2023).
\newblock Large language model for causal decision making.
\newblock {\em ArXiv}, abs/2312.17122.

\bibitem[Kasie et~al., 2017]{Kasie2017DecisionSS}
Kasie, F.~M., Bright, G., and Walker, A.~J. (2017).
\newblock Decision support systems in manufacturing: a survey and future trends.
\newblock {\em Journal of Modelling in Management}, 12:00--00.

\bibitem[Krantz and Kunreuther, 2007]{Krantz2007GoalsAP}
Krantz, D.~H. and Kunreuther, H. (2007).
\newblock Goals and plans in decision making.
\newblock {\em Judgment and Decision Making}.

\bibitem[Liu et~al., 2021]{Liu2021DecisionMakingTF}
Liu, Q., Li, X., Yuan, S., and Li, Z. (2021).
\newblock Decision-making technology for autonomous vehicles: Learning-based methods, applications and future outlook.
\newblock {\em 2021 IEEE International Intelligent Transportation Systems Conference (ITSC)}, pages 30--37.

\bibitem[Luger and Stubblefield, 1990]{Luger1990ArtificialIA}
Luger, G.~F. and Stubblefield, W.~A. (1990).
\newblock Artificial intelligence and the design of expert systems.

\bibitem[Meng et~al., 2021]{Meng2021OfflinePM}
Meng, L., Wen, M., Yang, Y., Le, C., Li, X., Zhang, W., Wen, Y., Zhang, H., Wang, J., and Xu, B. (2021).
\newblock Offline pre-trained multi-agent decision transformer: One big sequence model tackles all smac tasks.
\newblock {\em ArXiv}, abs/2112.02845.

\bibitem[OpenAI, 2024]{openai2024gpt4}
OpenAI (2024).
\newblock Gpt-4 technical report.

\bibitem[Ortolano, 1984]{Ortolano1984EnvironmentalPA}
Ortolano, L. (1984).
\newblock Environmental planning and decision making.

\bibitem[Ouyang et~al., 2022]{Ouyang2022TrainingLM}
Ouyang, L., Wu, J., Jiang, X., Almeida, D., Wainwright, C.~L., Mishkin, P., Zhang, C., Agarwal, S., Slama, K., Ray, A., Schulman, J., Hilton, J., Kelton, F., Miller, L.~E., Simens, M., Askell, A., Welinder, P., Christiano, P.~F., Leike, J., and Lowe, R.~J. (2022).
\newblock Training language models to follow instructions with human feedback.
\newblock {\em ArXiv}, abs/2203.02155.

\bibitem[Pannu and Student, 2015]{Pannu2015SurveyOE}
Pannu, A. and Student, M.~T. (2015).
\newblock Survey on expert system and its research areas.

\bibitem[Phillips-Wren and Jain, 2006]{Phillips-Wren}
Phillips-Wren, G. and Jain, L. (2006).
\newblock Artificial intelligence for decision making.
\newblock volume 4252, pages 531--536.

\bibitem[Radford et~al., 2019]{Radford2019LanguageMA}
Radford, A., Wu, J., Child, R., Luan, D., Amodei, D., and Sutskever, I. (2019).
\newblock Language models are unsupervised multitask learners.

\bibitem[Reed et~al., 2022]{Reed2022AGA}
Reed, S., Zolna, K., Parisotto, E., Colmenarejo, S.~G., Novikov, A., Barth-Maron, G., Gimenez, M., Sulsky, Y., Kay, J., Springenberg, J.~T., Eccles, T., Bruce, J., Razavi, A., Edwards, A.~D., Heess, N. M.~O., Chen, Y., Hadsell, R., Vinyals, O., Bordbar, M., and de~Freitas, N. (2022).
\newblock A generalist agent.
\newblock {\em Trans. Mach. Learn. Res.}, 2022.

\bibitem[Rozi{\`e}re et~al., 2023]{Rozire2023CodeLO}
Rozi{\`e}re, B., Gehring, J., Gloeckle, F., Sootla, S., Gat, I., Tan, X., Adi, Y., Liu, J., Remez, T., Rapin, J., Kozhevnikov, A., Evtimov, I., Bitton, J., Bhatt, M.~P., Ferrer, C.~C., Grattafiori, A., Xiong, W., D'efossez, A., Copet, J., Azhar, F., Touvron, H., Martin, L., Usunier, N., Scialom, T., and Synnaeve, G. (2023).
\newblock Code llama: Open foundation models for code.
\newblock {\em ArXiv}, abs/2308.12950.

\bibitem[Sharipbay et~al., 2024]{Sharipbay}
Sharipbay, A., Umutkulov, D., Bektemyssova, G., and Nisheva-Pavlova, M. (2024).
\newblock Analysis of the application of expert systems.
\newblock {\em Bulletin of the National Engineering Academy of the Republic of Kazakhstan}, 89:128--138.

\bibitem[Touvron et~al., 2023]{Touvron2023Llama2O}
Touvron, H., Martin, L., Stone, K.~R., Albert, P., Almahairi, A., Babaei, Y., Bashlykov, N., Batra, S., Bhargava, P., Bhosale, S., Bikel, D.~M., Blecher, L., Ferrer, C.~C., Chen, M., Cucurull, G., Esiobu, D., Fernandes, J., Fu, J., Fu, W., Fuller, B., Gao, C., Goswami, V., Goyal, N., Hartshorn, A.~S., Hosseini, S., Hou, R., Inan, H., Kardas, M., Kerkez, V., Khabsa, M., Kloumann, I.~M., Korenev, A.~V., Koura, P.~S., Lachaux, M.-A., Lavril, T., Lee, J., Liskovich, D., Lu, Y., Mao, Y., Martinet, X., Mihaylov, T., Mishra, P., Molybog, I., Nie, Y., Poulton, A., Reizenstein, J., Rungta, R., Saladi, K., Schelten, A., Silva, R., Smith, E.~M., Subramanian, R., Tan, X., Tang, B., Taylor, R., Williams, A., Kuan, J.~X., Xu, P., Yan, Z., Zarov, I., Zhang, Y., Fan, A., Kambadur, M., Narang, S., Rodriguez, A., Stojnic, R., Edunov, S., and Scialom, T. (2023).
\newblock Llama 2: Open foundation and fine-tuned chat models.
\newblock {\em ArXiv}, abs/2307.09288.

\bibitem[Vaswani et~al., 2017]{Vaswani2017AttentionIA}
Vaswani, A., Shazeer, N.~M., Parmar, N., Uszkoreit, J., Jones, L., Gomez, A.~N., Kaiser, L., and Polosukhin, I. (2017).
\newblock Attention is all you need.
\newblock In {\em Neural Information Processing Systems}.

\bibitem[Wei et~al., 2022]{Wei2022EmergentAO}
Wei, J., Tay, Y., Bommasani, R., Raffel, C., Zoph, B., Borgeaud, S., Yogatama, D., Bosma, M., Zhou, D., Metzler, D., hsin Chi, E.~H., Hashimoto, T., Vinyals, O., Liang, P., Dean, J., and Fedus, W. (2022).
\newblock Emergent abilities of large language models.
\newblock {\em ArXiv}, abs/2206.07682.

\bibitem[Wen et~al., 2022a]{Wen2022MultiAgentRL}
Wen, M., Kuba, J.~G., Lin, R., Zhang, W., Wen, Y., Wang, J., and Yang, Y. (2022a).
\newblock Multi-agent reinforcement learning is a sequence modeling problem.
\newblock {\em ArXiv}, abs/2205.14953.

\bibitem[Wen et~al., 2022b]{Wen2022OnRO}
Wen, Y., Wan, Z., Zhou, M., Hou, S., Cao, Z., Le, C., Chen, J., Tian, Z., Zhang, W., and Wang, J. (2022b).
\newblock On realization of intelligent decision-making in the real world: A foundation decision model perspective.
\newblock {\em ArXiv}, abs/2212.12669.

\bibitem[Yang et~al., 2023]{Yang2023FoundationMF}
Yang, S., Nachum, O., Du, Y., Wei, J., Abbeel, P., and Schuurmans, D. (2023).
\newblock Foundation models for decision making: Problems, methods, and opportunities.
\newblock {\em ArXiv}, abs/2303.04129.

\bibitem[Zhang, 2023]{10.24963/ijcai.2023/808}
Zhang, W. (2023).
\newblock Large decision models.
\newblock In {\em Proceedings of the Thirty-Second International Joint Conference on Artificial Intelligence}, IJCAI '23.

\end{thebibliography}


\end{document}